\documentclass{llncs}
\usepackage{llncsdoc}

\usepackage[center,justification=justified,singlelinecheck=true,indention=1em,margin=10pt]{caption}

\renewcommand{\phi}{\varphi}
\usepackage{color,graphicx}
\usepackage{amssymb}
\usepackage{amsmath}
\usepackage{array}
\usepackage{mathtools}
\usepackage{tabularx}
\mathtoolsset{showonlyrefs}
\newcounter{rowcount}
\title{Automated Reasoning in Deontic Logic\thanks{Work supported by DFG grants FU~263/15-1 and STO~421/5-1 'Ratiolog'.}}

\author{Ulrich Furbach\inst{1}, Claudia Schon\inst{1} and Frieder Stolzenburg\inst{2}}
\institute{
	Universit\"at Koblenz-Landau, 
    \email{\{uli,schon\}@uni-koblenz.de}
 \and
	Harz University of Applied Sciences,
   \email{fstolzenburg@hs-harz.de}
}

\begin{document}
\maketitle

\begin{abstract}
Deontic logic is a very well researched branch of mathematical logic and
philosophy. Various kinds of deontic logics are discussed for different
application domains like argumentation theory, legal reasoning, and acts in
multi-agent systems. In this paper, we show how standard deontic logic 
can be stepwise transformed into description logic and DL-clauses, such that it
can be processed by Hyper, a high performance theorem prover which uses a
hypertableau calculus. Two use cases, one from multi-agent research and one from
the development of normative system are investigated.
\end{abstract}

\section{Introduction} 
Deontic logic is a very well researched branch of mathematical logic and•
philosophy. Various kinds of deontic logics are discussed for different
application domains like argumentation theory, legal reasoning, and acts in
multi-agent systems \cite{handbook}. Recently there also is growing interest in
modelling human reasoning and testing the models with psychological findings.
Deontic logic is an obvious tool to this end, because norms and licenses in
human societies can be described easily with it. For example in
\cite{DBLP:journals/corr/FurbachS14} there is a discussion of some of these 
problems including  solutions with the help of deontic logic. There, the focus
is on using deontic logic for modelling certain effects, which occur in human
reasoning, e.g. the Wason selection task or Byrne's suppression task.

This paper concentrates on automated reasoning in standard deontic logic
(SDL). Instead of
implementing a reasoning system for this logic directly, we rather rely on
existing methods and systems. Taking into account that SDL is just the modal logic $\mathsf{K}$ with a seriality axiom, we show that
deontic logic can be translated into  description logic $\mathcal{ALC}$. The
latter can be transformed into so called DL-clauses, which is a special normal
form with clauses consisting of implications where the body is, as usual, a
conjunction of atoms and the head is a disjunction of literals. These literals
can be atoms or existential quantified expressions.

DL-clauses can be decided
by the first-order reasoning system Hyper \cite{WernhardPelzer}, which uses the
hypertableau calculus from \cite{DBLP:conf/jelia/BaumgartnerFN96}. In
Sections \ref{sec:modal}~and~\ref{hyper} we shortly depict this workflow, and in Section~\ref{sect:applications} we
demonstrate the use of our technique with the help of two problems from the
literature, one from multi-agent research and the other one from testing
normative systems. We choose these examples, because they hopefully document the
applicability of reasoning of SDL in various areas of AI research.
  
\section{Deontic Logic as Modal Logic $\mathsf{KD}$}\label{sec:modal}
We consider a simple modal logic which consists of propositional logic and the additional modal operators $\Box$ and $\Diamond$.
Semantics are given as possible world semantics, where the modal operators $\Box$ and $\Diamond$ are interpreted as quantifiers over possible worlds. Such a possible world is an assignment,
which assigns truth values to the propositional variables. An interpretation
connects different possible worlds by a (binary) reachability relation $R$. The
$\Box$-operator states that a formula has to hold in all reachable worlds. Hence
if $v$ and $w$ are  worlds, we have
\begin{equation*}
w \models \Box P  \hspace{2em} \textrm{iff} \hspace{2em} \forall v : R(w,v) \rightarrow v \models P
\end{equation*}

Standard deontic logic (SDL) is obtained from the well-known modal logic
$\mathsf{K}$ by adding the seriality axiom $\mathsf{D}$:
\[  \mathsf{D:} \quad \Box P \rightarrow \Diamond P \]
In this logic, the  $\Box$-operator is interpreted  as `it is obligatory that'
and the $\Diamond$ as `it is permitted that'. The $\Diamond$-operator can be
defined by the following equivalence:
\[ \Diamond P \equiv \neg\Box\neg P \]

The additional axiom $\mathsf{D}$: $\Box P \rightarrow
\Diamond P$ in SDL states that, if a formula  has to hold in all reachable
worlds, then there exists such a world. With the deontic reading of $\Box$ and
$\Diamond$ this means: Whenever the formula $P$ ought to be, then there
exists a world where it holds. In consequence, there is always a world, which is
ideal in the sense, that all the norms formulated by `the ought to
be'-operator hold.

SDL can be used in a natural way to describe knowledge about norms or licenses. The use of conditionals for expressing rules which should be considered as norms seems likely, but holds some subtle difficulties. If we want to express that \emph{if $P$ then $Q$} is a norm, an obvious solution would be to use 
\[\Box ( P \rightarrow  Q)\]
which reads \emph{it is obligatory that Q holds if P holds}. An alternative would be 
\[P \rightarrow \Box Q\]
meaning \emph{if P holds, it is obligatory that Q holds}.
In \cite{kutschera} there is a careful discussion which of these two
possibilities should be used for conditional norms. The first one has severe
disadvantages. The most obvious disadvantage is, that $P$ together with $\Box (
P \rightarrow  Q)$ does not imply $\Box Q$. This is why we prefer the latter 
method, where the $\Box$-operator is in the conclusion of the conditional. We
will come back to this point in Subsection~\ref{sec:consistency} where we consider
several formalization variants of the well-known problem of
contrary-to-duty-obligations. For a more detailed discussion of such aspects we
refer to \cite{gabbay2013handbook}.

\section{Automated Reasoning for Deontic Logic}\label{hyper}

Deontic logic is the logic of choice when formalizing knowledge about norms like
the representation of legal knowledge. However, there are only few automated
theorem provers specially dedicated for deontic logic and used by deontic logicians (see
\cite{Artosi94ked:a,Bassiliades:2011:MDR:2441484.2441486}). Nonetheless,
numerous approaches to translate modal logics into (decidable fragments
of) first-order predicate logics are stated in the literature. A nice overview
including many relevant references is given in \cite{SH13}. 

In this paper, we describe how to use the Hyper theorem prover
\cite{WernhardPelzer} to handle deontic logic knowledge bases. These knowledge
bases can be translated efficiently into description logic formulae.
Hyper is a theorem prover for first-order logic with equality. 
In addition to that, Hyper is a decision procedure for the description logic $\mathcal{SHIQ}$ \cite{cadesd}.

\begin{figure}[top]
\centering
\includegraphics[scale=0.4]{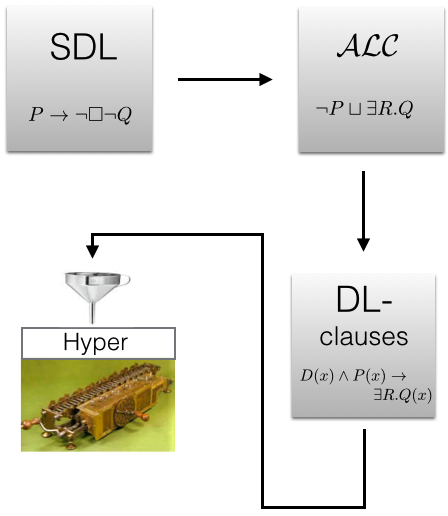}
\caption{From SDL to Hyper. Note that concept $D$ occurring in the DL-clauses is an auxiliary concept.}
\label{fig:uebrsicht}
\end{figure}

In Figure~\ref{fig:uebrsicht}, we depict the entire workflow from a given SDL
knowledge base to the final input into the Hyper theorem prover. In the
following, we describe these three steps in more detail.

\subsection{Transformation from Deontic Logic into $\mathcal{ALC}$}\label{subsec:translation}

First, we will show how to translate SDL knowledge bases into $\mathcal{ALC}$
knowledge bases. An $\mathcal{ALC}$ knowledge base consists of a TBox and an
ABox. The TBox (terminological box) gives information about concepts occurring
in the domain of interest and describes concept hierarchies. The ABox
(assertional box) introduces individuals and states, to which concepts the
individuals belong and how they are interconnected via relations called roles .
The ABox contains assertional knowledge and can be seen as the representation of
a state of the world. We do not give the syntax and semantics of
$\mathcal{ALC}$ here and refer the reader to \cite{DLhandbookCh2}.

There is a strong connection between
modal logic and the description logic $\mathcal{ALC}$.  As shown in
\cite{Schild91acorrespondence},  the description logic $\mathcal{ALC}$ is  a
notational variant of the modal logic $\mathsf{K}_{n}$. Therefore any formula
given in the modal logic $\mathsf{K}_{n}$ can be translated into an
$\mathcal{ALC}$ concept and vice versa.  Since we are only considering a modal
logic as opposed to a multimodal logic, we will omit the part of the translation
handling the multimodal part of the logic. 
Mapping $\phi$ translating from modal logic $\mathsf{K}$
formulae to $\mathcal{ALC}$ concepts is inductively defined as follows:
\begin{center}
\begin{tabular}{rcl}
$\phi(\top)$ 			& = &  $\top$\\
$\phi(\bot)$ 			& = &  $\bot$\\
$\phi(b)$ 				& = &  $b$\\
$\phi(\lnot c)$ 			& = & $\lnot \phi(c)$\\
$\phi(c \land d)$ 		& = & $\phi(c) \sqcap \phi(d)$\\
$\phi(c \lor d)$ 		& = & $\phi(c) \sqcup \phi(d)$\\
$\phi(\Box c)$ 			& = & $\forall r.\phi(c)$\\
$\phi(\Diamond c)$ 	& = & $\exists r.\phi(c)$
\end{tabular}
\end{center}
Note that the mapping $\phi$ is a one-to-one mapping.

Formulae given in SDL can be translated into $\mathcal{ALC}$ concepts using the above introduced $\phi$ mapping.
 For a normative system consisting of the set of deontic logic formulae
$\mathcal{N}= \lbrace F_{1},\ldots, F_{n}\rbrace$
the translation is defined as the conjunctive combination of the translation of all deontic logic formulae in $\mathcal{N}$:
\begin{equation}
\phi(\mathcal{N}) = \phi(F_{1}) \sqcap \ldots \sqcap \phi(F_{n})
\end{equation}

Note that $\phi(\mathcal{N})$ does not yet contain the translation of the seriality axiom. As shown in \cite{Klarman28052013} the seriality axiom can be translated into the following TBox:
\begin{equation*}
\mathcal{T}= \lbrace \top \sqsubseteq \exists r.\top\rbrace
\end{equation*}
with $r$ the atomic role introduced by the mapping $\phi$.

For our application, the result of the translation of a normative system $\mathcal{N}$ and the seriality axiom is an $\mathcal{ALC}$ knowledge base $\Phi(\mathcal{N})=(\mathcal{T},\mathcal{A})$,
where the TBox $\mathcal{T}$ consists of the translation of the seriality axiom and the ABox
$\mathcal{A}= \lbrace (\phi(\mathcal{N}))(a)\rbrace$ for a new individual $a$. In description logics performing a satisfiability test of a concept $C$ w.r.t. a TBox  is usually done by adding a new individual $a$ together with the ABox assertion $C(a)$. For the sake of simplicity, we do this construction already during the transformation of $\Phi$ by adding $(\phi(\mathcal{N}))(a)$ to the ABox.

An advantage of the translation of deontic logic formulae into an
$\mathcal{ALC}$ knowledge base is the existence of a TBox in $\mathcal{ALC}$. 
This makes it possible to add further axioms to the TBox. For example we can add
certain norms that we want to be satisfied in all reachable worlds into the
TBox.

\subsection{Translation from $\mathcal{ALC}$ into DL-Clauses}
Next we transform the $\mathcal{ALC}$ knowledge base into so called DL-clauses introduced in \cite{msh07optimizing} which represent the input format for the Hyper theorem prover.

DL-clauses are constructed from so called $\emph{atoms}$. 
An atom is of the form $b(s)$, $r(s,t)$, $\exists r.b(s)$ or $\exists r.\lnot
b(s)$ for $b$ an atomic concept and $s$ and $t$ individuals or variables.
They are universally quantified implications of the form 
\begin{equation}
\bigwedge_{i=1}^{m} u_{i} \rightarrow \bigvee_{j=1}^{n} v_{j}
\end{equation}
where the $u_{i}$ are atoms of the form $b(s)$ or $r(s,t)$ and the $v_j$ may be
arbitrary DL-clause atoms, i.e. including existential quantification, with $m,n
\geq 0$. 

Comparing the syntax of DL-clauses to the syntax of first order logic clauses written as implications, the first obvious difference is the absence of function symbols. The second difference is the fact, that in DL-clauses all atoms are constructed from unary or binary predicates. The most interesting difference however is the fact, that the head of a DL-clause is allowed to contain atoms of the form $\exists r.b(s)$.

The basic idea of the translation of an $\mathcal{ALC}$ knowledge base into DL-clauses is that the subsumption in a TBox assertion is interpreted as an implication from the left to the right side. Further concepts are translated to unary and roles to binary predicates. Depending on the structure of the assertion, auxiliary concepts are introduced.
For example the TBox axiom
\begin{equation}
d \sqsubseteq \exists r.b \sqcup \forall r.c
\end{equation}
corresponds to the following DL-clause
\begin{equation}
d(x) \land r(x,y) \rightarrow c(y) \lor \exists r.b(x) 
\end{equation}
For detailed definitions of both syntax and semantics of DL-clauses and the translation into DL-clauses, we refer the reader to \cite{msh07optimizing}. The translation preserves equivalence, avoids an exponential blowup by using a well-known structural transformation \cite{DBLP:journals/jsc/PlaistedG86} and can be computed in polynomial time.
In the following, for an $\mathcal{ALC}$ knowledge base $\mathcal{K} =
(\mathcal{T},\mathcal{A})$, the corresponding set of DL-clauses is denoted by $\omega(\mathcal{K})$.

\subsection{Reasoning Tasks}\label{sec:reasoning tasks}
With the help of Hyper, we can solve several interesting reasoning tasks:
\begin{itemize}
\item \textbf{Consistency checking of normative systems:} In practice, normative systems can be very large. Therefore it is not easy to see, if a given normative system is consistent. The Hyper theorem prover can be used to check consistency of a normative system $\mathcal{N}$. We first translate $\mathcal{N}$ into an $\mathcal{ALC}$ knowledge base $\Phi(\mathcal{N})$, then translate $\Phi(\mathcal{N})$ into the set $\omega(\Phi(\mathcal{N}))$ of DL-clauses. Then we can check the consistency of $\omega(\Phi(\mathcal{N}))$ using Hyper.
\item \textbf{Evaluation of normative systems:} Given several normative systems, we use Hyper to find out for which normative system guarantees a desired outcome is guaranteed. 
\item \textbf{Independence checking:} Given a normative system $\mathcal{N}$ and a formula $F$ representing a norm, we can check whether $F$ is independent from $\mathcal{N}$. If $F$ is independent from $\mathcal{N}$, then $F$ is not a logical consequence of $\mathcal{N}$.
\end{itemize}

In Section~\ref{sect:applications}, we will give detailed examples for those
tasks. Subsection~\ref{sec:consistency} gives an example for a consistency check
of a normative system and illustrates how the independence of a formula from a
normative system can be decided. In Subsection~\ref{sec:multi-agent}, we use an
example from multi-agent systems to show how to evaluate normative systems.

\section{Applications}\label{sect:applications}


The literature on deontic logic deals with numerous small but nonetheless
interesting examples. They are mostly used to show typical problems or special
features of the logic under consideration (cf.~\cite{gabbay2013handbook}). In
Subsection~\ref{sec:consistency}, we deal with one of these
examples. In Subsection~\ref{sec:multi-agent}, we formalize a `real-life'
problem from multi-agent research.

\subsection{Contrary-to-duty Obligations}
\label{sec:consistency}
Let us now consider consistency testing of normative systems and independence
checking. As an example, we examine the well-known problem of
\emph{contrary-to-duty obligations} introduced in \cite{chisolm}:

\begin{enumerate}
\renewcommand{\labelenumi}{(\arabic{enumi})}\item $a$ ought not steal.
\item $a$ steals.
\item If $a$ steals, he ought to be punished for stealing.
\item If $a$ does not steal, he ought not be punished for stealing.
\end{enumerate}
Table~\ref{tab:formalization} shows three different formalizations of this problem. Those formalizations are well-known from the literature \cite{2973603,mcnamara1999norms,McNamara2010,kutschera}:

\begin{table}[top]
\begin{center}
\begin{tabular*}{0.75\textwidth}{@{\extracolsep{\fill} } c|lll}
	&$\mathcal{N}_{1}$ 				& $\mathcal{N}_{2}$ 			& $\mathcal{N}_{3}$  \\
	\hline
{\refstepcounter{rowcount}\label{tab:line1}\hspace*{\tabcolsep}(\therowcount)\hspace*{\tabcolsep}}	&$\Box \lnot s$					& $\Box \lnot s$				& $\Box \lnot s$ 			\\
{\refstepcounter{rowcount}\label{tab:line2}\hspace*{\tabcolsep}(\therowcount)\hspace*{\tabcolsep}}	&$s$ 						& $s$ 						& $s$\\
{\refstepcounter{rowcount}\label{tab:line3}\hspace*{\tabcolsep}(\therowcount)\hspace*{\tabcolsep}}	&$s \rightarrow \Box p$ 			& $\Box(s \rightarrow p)$			& $s \rightarrow \Box p$\\
{\refstepcounter{rowcount}\label{tab:line4}\hspace*{\tabcolsep}(\therowcount)\hspace*{\tabcolsep}}	& $\Box (\lnot s \rightarrow \lnot p)$	& $\Box (\lnot s \rightarrow \lnot p)$	& $ \lnot s \rightarrow \Box \lnot p$	\\
\end{tabular*}
\end{center}
\caption{Formalizations of the \emph{contrary-to-duty obligation} introduced in \cite{chisolm}.}
\label{tab:formalization}
\end{table}

\subsubsection{Consistency Testing of Normative Systems}
The contrary-to-duty obligation formalized above is a very small example. In
practice, normative systems can be rather complex. This makes it difficult to
see if a normative system is consistent. We will show how to use the Hyper
theorem prover to check the consistency of a given normative system.

As an example, we consider formalization $\mathcal{N}_{1}$ given in Table~\ref{tab:formalization} which, according to \cite{kutschera}, is inconsistent. 
We will use Hyper to show this inconsistency. For this, we first translate normative system $\mathcal{N}_{1} $ into an $\mathcal{ALC}$ knowledge base $\Phi(\mathcal{N}_{1})$. Table~\ref{tab:translation2} shows $\phi(\mathcal{N}_{1} )$.
\begin{table}[top]
\begin{center}
\begin{tabular*}{0.4\textwidth}{@{\extracolsep{\fill} } ll  }
$\mathcal{N}_{1}$ (in Deontic Logic) & $\phi(\mathcal{N}_{1})$ \\
\hline
$\Box \lnot s$ 					& $\forall r. \lnot s$ 			\\
$s$ 							&  $s$ 					\\
$s \rightarrow \Box p$ 			& $\lnot s \sqcup \forall r.p$	\\
 $\Box (\lnot s \rightarrow \lnot p)$	& $\forall r.(s \sqcup \lnot p)$	\\
\end{tabular*}
\end{center}
\caption{Translation of the normative system $\mathcal{N}_{1}$  into $\phi(\mathcal{N}_{1}$).}
\label{tab:translation2}
\end{table}



To perform the satisfiability test, we transform the description logic representation $\Phi(\mathcal{N}_{1})$ into a set of DL-clauses $\omega(\Phi(\mathcal{N}_{1}))$.
Hyper constructs a hypertableau for $\omega(\Phi(\mathcal{N}_{1}))$. This hypertableau is closed and therefore we can conclude that $\mathcal{N}_{1}$ is inconsistent.

\subsubsection{Independence Checking}
Normative System $\mathcal{N}_{2}$ given in Table~\ref{tab:formalization} is
consistent. However it has another drawback: The different formulae in this
formalization are not independent from another. Formula~\eqref{tab:line3} is a
logical consequence of~\eqref{tab:line1}, because $\Box(s \rightarrow p) \equiv
\Box(\neg s \vee p)$ (definition of $\rightarrow$) which clearly is implied by
the (subsuming) formula \eqref{tab:line1}~$\Box \neg s$.
We can use Hyper to show this by transforming the problem into a satisfiability
test. For this, we remove formula~\eqref{tab:line3} from $\mathcal{N}_{2}$ and
add its negation $\lnot \Box (s \rightarrow p)$ to $\mathcal{N}_{2}$. If the
resulting normative system is inconsistent, we can conclude, that
formula~\eqref{tab:line3} is not independent from the other formulae in
$\mathcal{N}_{2}$.

The problem of independence of formulae given in a normative system is
interesting in practice as well. If an existing normative system is extended
with some new formulae, it is interesting to know, whether the new formulae are
independent from the original normative system. This can be checked
automatically using Hyper as described above.

In the same way, we can show, that formula~\eqref{tab:line4} is not independent
from $\mathcal{N}_{3}$. Note that only this normative system is both consistent and
represents all conditionals carefully, i.e. with formulae of the form $P
\rightarrow \Box Q$ (cf.~Section~\ref{sec:modal}). Only for this formalization
we have: If $a$ steals in the actual world, $a$ will be punished in the
corresponding reachable ideal world. 
%

\subsection{An Example from Multi-agent Systems}
\label{sec:multi-agent}

In multi-agent systems, there is  a relatively new area of research, namely the formalization of `robot ethics'. It aims at defining formal rules for  the behavior of  agents and to prove certain properties. As an  example  consider Asimov's laws, which aim at regulating the relation between robots and humans. In \cite{journals/expert/BringsjordAB06} the authors depict  a small example of two surgery robots obeying ethical codes concerning their work. These codes are expressed by means of MADL, which is an extension of standard deontic logic with two operators.
In \cite{Murakami04utilitariandeontic} an axiomatization of MADL is given. Further it is asserted, that MADL is not essentially different from standard deontic logic. This is why we use SDL to model the example.

\subsubsection{Formalization in SDL}\label{sec:formalizationindeonticlogic}
In our example, there are two robots $\mathit{ag1}$ and $\mathit{ag2}$ in a
hospital. For sake of simplicity, each robot can perform one specific action:
$\mathit{ag1}$ can terminate a person's life support and $\mathit{ag2}$ can
delay the delivery of pain medication.  In \cite{journals/expert/BringsjordAB06}
four different ethical codes $\mathit{J}$, $\mathit{J}^{\star}$, $\mathit{O}$
and $\mathit{O^{\star}}$ are considered:
\begin{itemize}
\item ``If ethical code $\mathit{J}$ holds, then robot $\mathit{ag1}$ ought to take care, that life support is terminated.'' This is formalized as:
\begin{equation}
\mathit{J}\rightarrow \Box  \mathit{act(ag1, term)}
\end{equation}
\item ``If ethical code $\mathit{J^{\star}}$ holds, then code $\mathit{J}$ holds, and robot $\mathit{ag2}$ ought to take care, that the delivery of pain medication is delayed.'' This is formalized as:
\begin{equation}
 \mathit{J^{\star}}\rightarrow \mathit{J} \land \mathit{J^{\star}} \rightarrow \Box  \mathit{act(ag2, delay)}
 \end{equation}
\item ``If ethical code $\mathit{O}$ holds, then robot $\mathit{ag2}$ ought to take care, that delivery of pain medication is not delayed.'' This is formalized as:
\begin{equation}
\mathit{O}\rightarrow \Box \lnot  \mathit{act(ag2, delay)}
\end{equation}
\item ``If ethical code $\mathit{O^{\star}}$ holds, then code $\mathit{O}$ holds, and robot $\mathit{ag1}$ ought to take care, that life support is not terminated.'' This is formalized as:
\begin{equation}
\mathit{O^{\star}}\rightarrow \mathit{O} \land \mathit{O^{\star}} \rightarrow \Box \lnot  \mathit{act(ag1, term)}
\end{equation}
\end{itemize}
Further we give a slightly modified version of the evaluation of the robot's acts given in \cite{journals/expert/BringsjordAB06}, where $(+!!)$ describes the most and $(-!!)$ the least desired outcome. Note that terms like $(+!!)$ are just propositional atomic formulae here.
\mathtoolsset{showonlyrefs=false}
\begin{align}
\mathit{act(ag1, term)} \land \phantom{\lnot} \mathit{act(ag2, delay)} & \rightarrow (-!!)\label{equ:eval1}\\
\mathit{act(ag1, term)} \land \lnot \mathit{act(ag2, delay)} & \rightarrow (-!)\label{equ:eval2}\\
\lnot \mathit{act(ag1, term)} \land \phantom{\lnot} \mathit{act(ag2, delay)} & \rightarrow (-)\label{equ:eval3}\\
\lnot \mathit{act(ag1, term)} \land \lnot \mathit{act(ag2, delay)} & \rightarrow (+!!)\label{equ:eval4}
\end{align}
\mathtoolsset{showonlyrefs=true}
These formulae evaluate the outcome of the robots' actions. It makes sense to assume, that this evaluation is effective in all reachable worlds. This is why we add formulae stating that formulae~\eqref{equ:eval1}--\eqref{equ:eval4} hold in all reachable worlds. For example, for \eqref{equ:eval1} we add:
\begin{equation}
\Box (\mathit{act(ag1, term)} \land \mathit{act(ag2, delay)} \rightarrow (-!!))\label{equ:allworlds}
\end{equation}
Since our example does not include nested modal operators, the formulae of the form \eqref{equ:allworlds} are sufficient to spread the evaluation formulae to all reachable worlds.
The normative system $\mathcal{N}$ formalizing this example consists of the formalization of the four ethical codes and the formulae for the evaluation of the robots actions.

\paragraph{Reduction to a Satisfiability Test}
A possible query would be to ask, if the most desirable outcome $(+!!)$ will come to pass, if ethical code $O^{\star}$ is operative. This query can be translated into a satisfiability test: If 
\begin{equation}
\mathcal{N} \land \mathit{O^{\star}} \land \Diamond \lnot (+!!)
\end{equation}
 is unsatisfiable, then ethical code $O^{\star}$ ensures outcome $(+!!)$.
 
\subsubsection{Translation into Description Logic}\label{sec:translationintodl}
As described in Section~\ref{subsec:translation}, we translate normative system $\mathcal{N}$ given in the previous section into an $\mathcal{ALC}$ knowledge base $\Phi(\mathcal{N})=(\mathcal{T},\mathcal{A})$. Table~\ref{tab:translation3} shows the result of translating $\mathcal{N}$ into $\phi(\mathcal{N})$.

\begin{table}[top]
\begin{scriptsize}
\begin{center}
\begin{tabular}{l | l}
\textbf{Deontic Logic} & $\mathcal{ALC}$ \\
\hline\\[-2ex]
%
\hspace{-.8ex}$\begin{aligned}								
	J 		&\rightarrow \Box \mathit{act}(\mathit{ag1},\mathit{term})\\
	J^{\star} 	&\rightarrow J \land J^{\star} \rightarrow \Box \mathit{act}(\mathit{ag2},delay)\\
	O 		&{}\rightarrow \Box \lnot \mathit{act}(\mathit{ag2},\mathit{delay})\\
	O^{\star} &\rightarrow O \land O^{\star} \rightarrow \Box \lnot \mathit{act}(\mathit{ag1},term)
\end{aligned}$ &	
\hspace{-.8ex}$\begin{aligned}								
	\ &\lnot J \sqcup \forall r. \mathit{act}(\mathit{ag1},term)\\
	&(\lnot J^{\star} \sqcup J) \sqcap ( \lnot J^{\star} \sqcup \forall r. \mathit{act}(\mathit{ag2},delay))\\
	&\lnot O \sqcup \forall r. \lnot \mathit{act}(\mathit{ag2},delay)\\
	&(\lnot O^{\star} \sqcup O) \sqcap ( \lnot O^{\star} \sqcup \forall r. \lnot \mathit{act}(\mathit{ag1},term))
\end{aligned}$\\\\[-2ex]
\hspace{-.8ex}$\begin{aligned}								
	\mathit{act(ag1, term)} \land \phantom{\lnot} \mathit{act(ag2, delay)} 		& \rightarrow (-!!)\\
	\mathit{act(ag1, term)} \land \lnot \mathit{act(ag2, delay)}  	& \rightarrow (-!)\\
	\lnot \mathit{act(ag1, term)} \land \phantom{\lnot}\mathit{act(ag2, delay)}  	& \rightarrow (-)\ \\
	\lnot \mathit{act(ag1, term)} \land \lnot \mathit{act(ag2, delay)}& \rightarrow (+!!)
\end{aligned}$ &
\hspace{-.8ex}$\begin{aligned}								
 \lnot (\mathit{act(ag1, term)} \sqcap \phantom{\lnot}\mathit{act(ag2, delay)}) 			&\sqcup (-!!)\\
 \lnot (\mathit{act(ag1, term)} \sqcap \lnot \mathit{act(ag2, delay)})   	&\sqcup (-!)\\
 \lnot (\lnot \mathit{act(ag1, term)} \sqcap \phantom{\lnot}\mathit{act(ag2, delay)})  	&\sqcup (-)\\
 \ \lnot (\lnot \mathit{act(ag1, term)} \sqcap \lnot \mathit{act(ag2, delay)}) &\sqcup (+!!)
\end{aligned}$ 
\end{tabular}
\end{center}
\end{scriptsize}
\caption{Translation of the normative system $\mathcal{N}$ into $\phi(\mathcal{N})$.}
\label{tab:translation3}
\end{table}

We further add the following two assertions to the ABox $\mathcal{A}$:
\begin{align}
O^{\star}(a)\label{equ:abox1}\\
\exists r. \lnot (+!!)(a)\label{equ:abox2}
\end{align}
Next we translate the knowledge base into DL-clauses and use Hyper to test the satisfiability of the resulting set of DL-clauses. 
Using further satisfiability tests, we can show, that ethical codes $J$, $J^{\star}$ or $O$ are not sufficient to guarantee the most desired outcome $(+!!)$.

\subsubsection{Formalization in Description Logic using a TBox}
In the formalization given in the previous subsection, we added formulae stating
that the evaluation of the agents' actions holds in all worlds, which are
reachable in one step, see~\eqref{equ:allworlds} for an example. In our case it
is sufficient to add formulae of the form~\eqref{equ:allworlds} because the
formalization does not include nested modal operators. In general it is
desirable to express that those formulae hold in \emph{all} reachable worlds including worlds reachable in more than one step. However
this would mean to either add infinitely many formulae or to use a universal
modality, i.e. the reflexive-transitive closure of the respective simple
modality. 

In description logics we can use a more elegant way to formalize that all worlds are supposed to fulfill certain formulae. Description logic knowledge bases contain a TBox including the terminological knowledge. Every individual is supposed to fulfill the assertions given in the TBox. Hence, we can  add the formulae stating the evaluation of the agents' actions into the TBox. 
For this, we reformulate implication ($\rightarrow$) by subsumption ($\sqsubseteq$).
We model the deontic logic formulae given in Table~\ref{tab:translation3} by the following TBox $\mathcal{T}$:
\begin{align}
\top & \sqsubseteq \exists r.\top\\
J & \sqsubseteq  \forall r. \mathit{act}(\mathit{ag1},\mathit{term})\\
J^{\star} & \sqsubseteq  J \\
J^{\star} & \sqsubseteq  \forall r. \mathit{act}(\mathit{ag2},delay)\\
O & \sqsubseteq \forall r. \lnot \mathit{act}(\mathit{ag2},\mathit{delay})\\
O^{\star} & \sqsubseteq  O\\
O^{\star} & \sqsubseteq\forall r. \lnot \mathit{act}(\mathit{ag}1,\mathit{term})\\
\mathit{act(ag1, term)} \sqcap \phantom{\lnot}\mathit{act(ag2, delay)} & \sqsubseteq (-!!)\\
\mathit{act(ag1, term)} \sqcap \lnot \mathit{act(ag2, delay)}  & \sqsubseteq (-!)\\
\lnot \mathit{act(ag1, term)} \sqcap \phantom{\lnot}\mathit{act(ag2, delay)}  & \sqsubseteq (-)\\
\lnot \mathit{act(ag1, term)} \sqcap \lnot \mathit{act(ag2, delay)}  & \sqsubseteq (+!!)
\end{align}
\paragraph{Reduction to a Satisfiability Test} Like in the previous section, we now want to know, if the most desirable outcome $(+!!)$ will come to pass, if ethical code $O^{\star}$ is operative. We perform this test by checking the satisfiability of the description logic knowledge base $\mathcal{K}=(\mathcal{T},\mathcal{A})$, with $\mathcal{T}$ as given above and $\mathcal{A}$ given as:
\begin{equation}
\mathcal{A}=  \lbrace O^{\star}(a), \exists r.\lnot (+!!)(a) \rbrace
\end{equation}
If this knowledge base is unsatisfiable, we can conclude, that $(+!!)$ will come to pass, if $O^{\star}$ is operative.
Again we can perform this satisfiability test, by translating the TBox and the ABox into DL-clauses and using Hyper to check the satisfiability.
We obtain the desired result, namely that (only) ethical code $O^\star$ leads to the most desirable behavior ($+!!$).

\subsection{Experiments}
We formalized the examples introduced in this section and tested it with the Hyper theorem prover as described above. Since all formalizations are available in $\mathcal{ALC}$, we used the description logic reasoner Pellet \cite{pellet-websem5} to show the unsatisfiability of the formalizations as well. Table~\ref{tab:experiment1} shows the results of our experiments. In the first column we see the time in seconds the two reasoners needed to show the unsatisfiability of the formalization of the example from multi-agent systems. For Hyper we give two different numbers. The first number is the time Hyper needs to show the unsatisfiability given the set of DL-clauses. In addition to that the second number contains the time needed to transform the $\mathcal{ALC}$ knowledge base into DL-clauses. The second column gives the runtimes for the example from multi-agent systems using the formalization with a TBox. And in the last column we present the runtimes for the consistency test of normative system $\mathcal{N}_{1}$ from the example on contrary-to-duty obligations.

\begin{table}[top]
\begin{center}
\begin{tabularx}{\textwidth}{l|p{11em}|p{11em}|p{11em}}
			& \textbf{Multi-agent Systems}				& \textbf{Multi-agent Systems (with TBox)}			& \textbf{Contrary-to-duty Obligations}\\
\hline
\textbf{Pellet} 	& 2.548			&  2.468		& 2.31\\
\hline
\textbf{Hyper} 	& 0.048 / 2.596 	& 0.048 / 2.102	& 0.03 / 1.749
\end{tabularx}
\end{center}
\caption{Time in seconds Pellet needed to show the unsatisfiability of the introduced examples. Time in seconds Hyper needed to show the unsatisfiability of the DL-clauses for the examples (the second number includes the translation into DL-clauses).}
\label{tab:experiment1}
\end{table}

For the examples we considered, the runtimes of Pellet and Hyper are
comparable.
Further investigation and comparison with other modal and/or description logic reasoning tools is required and subject of future work.
In order to use Hyper to perform the satisfiability tests, we first
have to translate the examples into DL-clauses. Our experiments show, that this
translation is not harmful. 

\section{Conclusion}

In this paper, we have demonstrated that by means of deontic logic complex
normative systems can be formalized easily. These formalizations can be checked
effectively with respect to consistency and independence from additional
formulae. For normative systems described with deontic logic, there is a
one-to-one translation into description logic formulae. These formula can be
checked automatically by automated theorem provers, which is in our case Hyper.

We are aware that deontic logic has several limitations. This is why future work aims at using more recent formalisms. 
For example we want to apply deontic logic in the context of natural-language
question-answering systems. There the normative knowledge in large databases
often leads to inconsistencies, which motivates us to consider combinations of deontic with
defeasible logic. 

\bibliographystyle{abbrv}
\bibliography{wissen}
\end{document}